\documentclass[10pt,twocolumn,letterpaper]{article}

\usepackage{cvpr}              

\usepackage[dvipsnames]{xcolor}

\definecolor{cvprblue}{rgb}{0.21,0.49,0.74}
\usepackage[pagebackref,breaklinks,colorlinks,citecolor=cvprblue]{hyperref}

\usepackage{graphicx}
\usepackage{stfloats}
\usepackage{multirow}
\usepackage{amssymb}
\usepackage{bbm}
\usepackage{href-ul}

\def\paperID{4038} 
\def\confName{CVPR}
\def\confYear{2024}

\title{StyleDyRF: Zero-shot 4D Style Transfer for Dynamic Neural Radiance Fields}

\newcommand*{\affaddr}[1]{#1} 
\newcommand*{\affmark}[1][*]{\textsuperscript{#1}}
\newcommand*{\email}[1]{\texttt{#1}}

\author{
Hongbin Xu\affmark[1], Weitao Chen\affmark[2], Feng Xiao\affmark[1], Baigui Sun\affmark[2], Wenxiong Kang\affmark[1]\\
\affaddr{\affmark[1]South China University of Technology}\\
\affaddr{\affmark[2]Alibaba Group}\\
\email{hongbinxu1013@gmail.com}
}

\begin{document}
\maketitle


\begin{abstract}

4D style transfer aims at transferring arbitrary visual style to the synthesized novel views of a dynamic 4D scene with varying viewpoints and times.
Existing efforts on 3D style transfer can effectively combine the visual features of style images and neural radiance fields (NeRF) but fail to handle the 4D dynamic scenes limited by the static scene assumption.
Consequently, we aim to handle the novel challenging problem of 4D style transfer for the first time, which further requires the consistency of stylized results on dynamic objects.
In this paper, we introduce StyleDyRF\footnote{Our code and model weights are released \href{https://github.com/ToughStoneX/StyleDyRF}{here}.}, a method that represents the 4D feature space by deforming a canonical feature volume and learns a linear style transformation matrix on the feature volume in a data-driven fashion.
To obtain the canonical feature volume, the rays at each time step are deformed with the geometric prior of a pre-trained dynamic NeRF to render the feature map under the supervision of pre-trained visual encoders.
With the content and style cues in the canonical feature volume and the style image, we can learn the style transformation matrix from their covariance matrices with lightweight neural networks.
The learned style transformation matrix can reflect a direct matching of feature covariance from the content volume to the given style pattern, in analogy with the optimization of the Gram matrix in traditional 2D neural style transfer.
The experimental results show that our method not only renders 4D photorealistic style transfer results in a zero-shot manner but also outperforms existing methods in terms of visual quality and consistency.

\end{abstract}    
\section{Introduction}


Given a reference style image, neural style transfer on 3D scene aims to render novel views which not only contain the target style but also hold consistency across different viewpoints in 3D scene.
The substantial prerequisite of 3D style transfer is the advances in 2D image-based neural style transfer \cite{gatys2016image,huang2017arbitrary,li2017universal,li2019learning} that allows transferring arbitrary styles in a zero-shot manner.
With the flourishing of 3D implicit representation of neural radiance fields (NeRF), many efforts have been conducted to combine the NeRF representation with neural style transfer techniques.
These prior studies \cite{huang2021learning,huang2022stylizednerf,mu20223d,liu2023stylerf} have proved that 3D style transfer should optimize novel view synthesis and style transfer jointly instead of naively stacking these components together to avoid the problem of multi-view inconsistency or poor stylization quality.

Nevertheless, all these approaches assume a static 3D scene without moving objects.
Despite their resounding success on static scenes, existing techniques on 3D style transfer are still deficient in handling dynamic scenes.
Consequently, in this paper, we first consider the novel challenge of \textbf{zero-shot 4D style transfer} (Fig. \ref{fig:motivation}) by relaxing the static assumption in 3D style transfer to 4D dynamic conditions, made of both static and moving/deforming objects.
Specifically, the imperative requirements of this task can be summarized as follows:
1) \textbf{Zero-shot generalization} towards arbitrary styles; 
2) \textbf{Multi-view consistency} of rendered stylized novel views;
3) \textbf{Cross-time consistency} of rendered stylized novel views;


Before elaborating on the detailed solution of this work, we would like to first revisit the typical pipeline of existing 3D zero-shot style transfer methods \cite{chiang2022stylizing,liu2023stylerf} to explore the bottleneck.
The typical pipeline contains 2 key ingredients: 
1) \emph{Explicit feature volume}:
As a radiance field can effectively capture the geometry and color information, the extension to feature space can endow expressive visual feature representation with 3D prior for great generalization performance.
2) \emph{Style transformation on the feature volume}:
The style transformation on the 3D-aware feature space can effectively maintain the multi-view consistency of the predicted stylized results.

In dynamic scenes, the explicit 3D feature volume suffers from the ambiguity of correspondences on moving objects, leading to an inaccurate 3D feature space.
Furthermore, the existing 3D style transformations vary with the changing 3D scenes over time, ignoring the global coherence across the time sequence.
In consequence, the core problem lies in: \emph{How to endow these 2 key components with cross-time consistency without losing their capability of generalization and multi-view consistency?}

In this paper, we propose \textbf{StyleDyRF} to resolve the challenging zero-shot 4D style transfer problem.
According to the aforementioned discussion, our StyleDyRF can be decomposed into 2 innovative designs:
1) \textbf{Canonical Feature Volume} (\textbf{CFV}):
To model the 4D temporal feature volume in the dynamic scene, our core idea is to decompose the feature radiance field into a canonical 3D feature volume and a deformation network based on the canonical space assumption \cite{pumarola2021d}.
The deformation network learns to map the 4D point $(x,y,z,t)$ at each time to the unified 3D canonical space, and sample feature by interpolating the canonical feature volume to render the 2D feature map at time $t$.
2) \textbf{Canonical Style Transformation} (\textbf{CST}):
The Whitening\&Coloring Transformation (WCT) \cite{li2017universal} in 2D style transfer reflects a direct matching of feature covariance from content to style.  
To ensure the temporal consistency of style transformation in 4D space, we attempt to learn the transformation matrix in a data-driven manner \cite{li2019learning}.
Specifically, we estimate the whitening and coloring transformation matrix with a light-weight 3D CNN on the global canonical feature volume and a 2D CNN on style images respectively.
Furthermore, for photorealistic 4D style transfer, we also present a post-processing module that can filter the distortions on the stylized novel views with the guidance of the volume-rendered images from the backbone NeRF.

The contribution of this work is summarized as follows:
1) We first investigate the novel challenge of zero-shot 4D style transfer, which has the following requirements: zero-shot generalization to arbitrary styles, multi-view consistency, and cross-time consistency. 
2) We introduce StyleDyRF, a novel zero-shot 4D style transfer framework that can effectively synthesize stylized novel views within the canonical feature space of a radiance field.
3) Extensive experiments show that our StyleDyRF can achieve superior performance in zero-shot 4D style transfer.
\section{Related Work}

\noindent\textbf{2D Style Transfer.}
Neural style transfer aims to synthesize a new image that can preserve the content structure of one image and meantime contain the style pattern of another.
The profound work \cite{gatys2016image} treats the style transfer as an iterative optimization task to match the correlations (Gram matrix) on the intermediate feature maps of pre-trained CNNs between images.
To speed up rendering, many followers \cite{deng2020arbitrary,huang2017arbitrary,johnson2016perceptual,li2019learning,li2017universal,liu2021adaattn,park2019arbitrary,sheng2018avatar,wu2021styleformer} apply feature transformations on the feature map extracted by pre-trained CNNs to approximate the time-consuming optimization process.
The feature transformations can be achieved by matching the mean and variance of features \cite{huang2017arbitrary}, matrix operation of whitening/coloring transformation \cite{li2017universal,li2019learning}, self-attention transformation \cite{deng2020arbitrary,liu2021adaattn,park2019arbitrary}, etc.
Video style transfer techniques extend previous progress in style transfer to videos by applying temporal constraints \cite{chen2017coherent,huang2017real,ruder2018artistic,wang2020consistent} across adjacent video frames.
Though they can generate smooth videos, the missing awareness of underlying 3D geometry limits their applications in rendering consistent frames in arbitrary novel views.

\noindent\textbf{3D Style Transfer.}
3D style transfer is a burgeoning field aiming to generate stylized content with multi-view consistency.
A simple solution is to combine the existing 2D style transfer techniques with neural radiance fields (NeRFs) \cite{mildenhall2021nerf,fridovich2022plenoxels,sun2022direct,chen2022tensorf,pumarola2021d,park2021nerfies,fang2022fast,liu2023robust} by first rendering novel images and then applying style transformation to the images.
Recent studies \cite{huang2021learning,chiang2022stylizing,huang2022stylizednerf,mu20223d,nguyen2022snerf,zhang2022arf,hollein2022stylemesh,liu2023stylerf,chen2022upst,fan2022unified} have shown that the naive combination of NeRF and 2D style transfer may result in unexpected multi-view inconsistency and poor stylization quality.
Huang et al \cite{huang2021learning} first handle the 3D style transfer by back-projecting image features to construct a point cloud and then perform style transformations on the point cloud.
StylizedNeRF \cite{huang2022stylizednerf} proposes a novel mutual learning framework with a 2D image stylization network and NeRF for style optimization.
To embed the generalization ability into the 3D style transfer model, \cite{chiang2022stylizing} utilize a hypernetwork comprised of MLPs to transfer the style information into the scene representation.
StyleRF \cite{liu2023stylerf} achieves state-of-the-art performance in 3D style transfer by transforming the content and style feature in the 3D space.
However, all these prior works concentrate on tackling the 3D style transfer problem, neglecting the 4D scenes with dynamic objects.
The dynamic motion in 4D scenes may corrupt the static assumption in all these prior works and thus deteriorate the rendering performance. 
In this paper, we aim to handle the 4D style transfer in a zero-shot manner, and our StyleDyRF can effectively handle the dynamic scenes and provide high-quality stylization on novel view synthesis.



\section{Method}
\label{sec:method}

\begin{figure*}[t]
    \centering
    \includegraphics[width=0.9\textwidth]{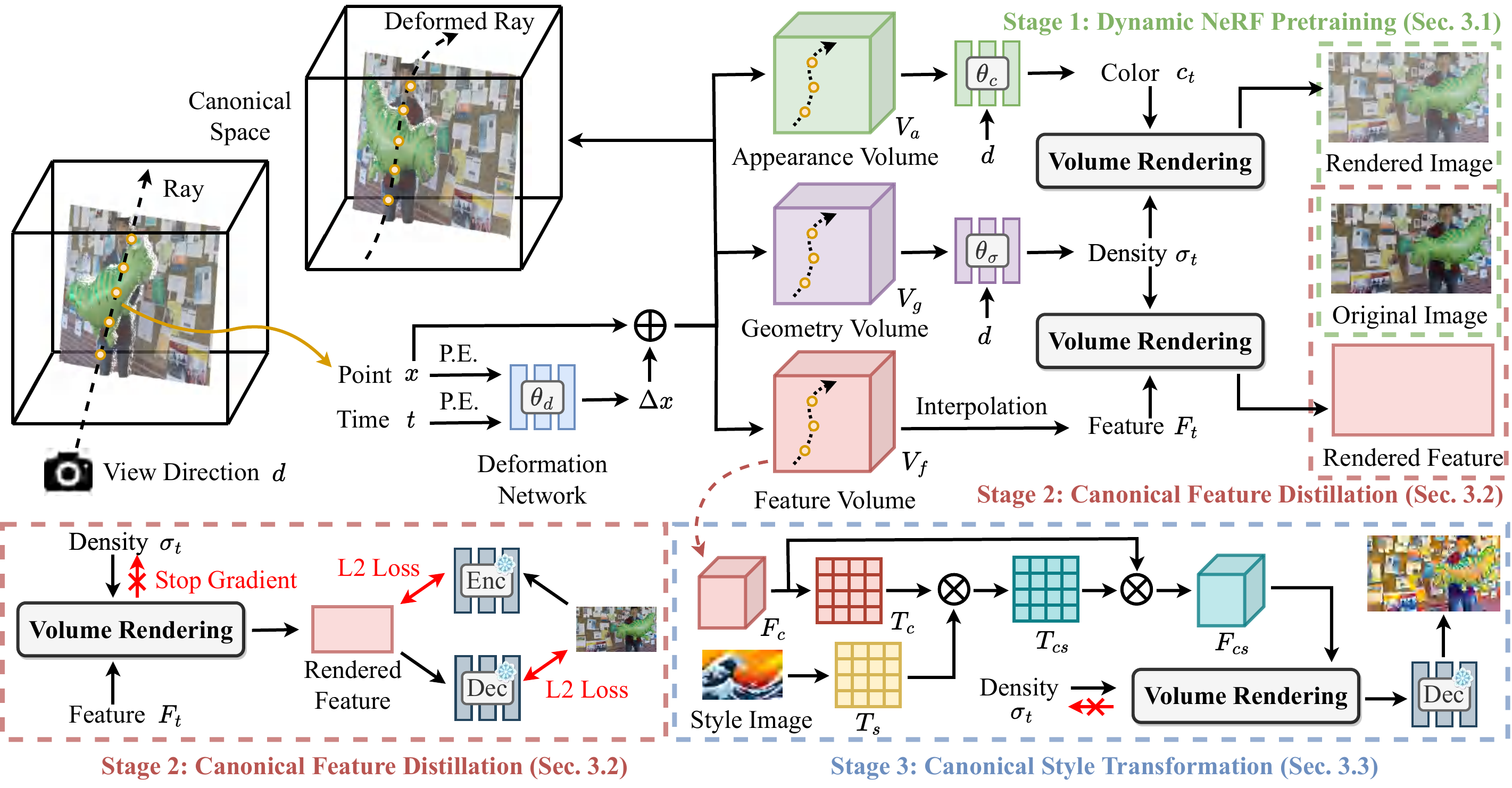}
    \vspace{-5pt}
    \caption{\textbf{The overview framework of StyleDyRF.}}
    \vspace{-10pt}
    \label{fig:framework}
\end{figure*}

The overview framework of our proposed StyleDyRF is provided in Fig. \ref{fig:framework}, which contains multiple stages for training. 
In the NeRF pre-training stage (Sec. \ref{sec:method:preliminary}), a dynamic NeRF is trained via volume rendering the images.
In the Canonical Feature Distillation stage (Sec. \ref{sec:method:canonical_feature_distillation}), the geometry branch of the pre-trained dynamic NeRF is used to guide the rendering from the canonical feature volume.
In the Canonical Style Transformation stage (Sec. \ref{sec:method:canonical_style_transformation}), we learn a linear style transformation matrix with lightweight networks from the canonical feature volume and the style image.

\subsection{Preliminary of Dynamic NeRF Pretraining}
\label{sec:method:preliminary}

We extend the basic NeRF \cite{pumarola2021d} to a dynamic one by attaching a deformation network comprised of MLPs parameterized by $\theta_d$.
The deformation network can estimate the deformation on the sampled point  $x \in \mathbb{R}^{3}$ at time $t \in \mathbb{R}$:
\begin{small}
\begin{equation}
    \setlength\abovedisplayskip{6pt}
    \setlength\belowdisplayskip{6pt}
    {\Delta x}_t = \text{MLP}_{\theta_d} (x, t)
    \label{eq1}
\end{equation}
\end{small}
Then we can project point $x$ back to the canonical space and sample the corresponding feature from the canonical appearance volume $V_a \in \mathbb{R}^{N_x \times N_y \times N_z \times C_a}$ and canonical geometry volume $V_g \in \mathbb{R}^{N_x \times N_y \times N_z \times C_g}$:
\begin{small}
\begin{equation}
    \setlength\abovedisplayskip{6pt}
    \setlength\belowdisplayskip{6pt}
    (v_a, v_g) = (V_a [ x+ {\Delta x}_t ], V_g [ x+{\Delta x}_t ] )
    \label{eq2}
\end{equation}
\end{small}
where $[ x+ {\Delta x}_t ]$ means the trilinear interpolation on the volume space based on the position of $x+ {\Delta x}_t$.
$v_a \in \mathbb{R}^{C_a}$ and $v_g \in \mathbb{R}^{C_g}$ are the sampled features.
After that, the density $\sigma_{t} \in \mathbb{R}$ and color $c_{t} \in \mathbb{R}^3$ can be computed via:
\begin{small}
\begin{equation}
    \setlength\abovedisplayskip{6pt}
    \setlength\belowdisplayskip{6pt}
    (c_t, \sigma_t) = (\text{MLP}_{\theta_{c}} (v_a, d, t), \text{MLP}_{\theta_{g}} (v_g, d, t))
    \label{eq3}
\end{equation}
\end{small}
where $d \in \mathbb{R}^3$ is the view direction.
The color $c_t$ and density $\sigma_t$ are respectively predicted with MLPs paramterized by $\theta_c$ and $\theta_{\sigma}$.

Then, the approximated volume rendering along the ray $r$ is used to calculate the pixel color:
\begin{small}
\begin{gather}
    \setlength\abovedisplayskip{6pt}
    \setlength\belowdisplayskip{6pt}
    \hat{I}_t (r) = \sum_{i=1}^{N_s} \omega_{t,i} c_{t,i} \label{eq4} \\
    \text{where} \;\; \omega_{t,i} = \text{exp} \left( -\sum_{j=1}^{i-1} \sigma_{t,j} \delta_j \right) \left( 1 - \text{exp} (-\sigma_{t,i} \delta_i )\right) \label{eq5}
\end{gather}
\end{small}
where $N_s$ points are sampled along the ray $r$.
$\sigma_{t,i}$ and $c_{t,i}$ are respectively the density and color predicted by Eq. \ref{eq3}.
$\delta_i$ is the distance between two adjacent points sampled on the ray.
Then we can optimize the radiance fields by minimizing the following photometric loss:
\begin{small}
\begin{equation}
    \setlength\abovedisplayskip{6pt}
    \setlength\belowdisplayskip{6pt}
    L_{\text{pho}} = \| \hat{I}_t (r) - I_t (r) \|_2^2
    \label{eq6}
\end{equation}
\end{small}
where $I_t$ is the ground truth image at time $t$.

\subsection{Canonical Feature Distillation}
\label{sec:method:canonical_feature_distillation}

To model the entire scene with deep features, we aim to find a volumetric representation to capture the representative features distilled from pre-trained CNNs like VGG \cite{simonyan2014very}.
In previous works, StyleRF \cite{liu2023stylerf} builds a feature volume in 3D space and optimizes it by rendering the 2D feature maps to approximate the ones extracted from given views.
However, such a 3D feature volume is insufficient to meet the need for 4D style transfer in dynamic scenes, because of the motion ambiguity caused by static assumption.
A naive solution is to directly extend the 3D feature volume to a 4D one, it is too memory-intensive to be embedded into the training framework.
Furthermore, the temporal consistency among different times should also be considered because the moving points should also be transformed consistently.
Consequently, we propose Canonical Feature Volume (CFV) to extend the canonical space assumption to the 3D feature space.
It can provide a compact 4D representation in the canonical 3D space and preserve the equivalent transformation across time simultaneously.

With the deformation network pre-trained in Sec. \ref{sec:method:preliminary}, we can sample the feature from the canonical feature volume $V_f \in \mathbb{R}^{N_x \times N_y \times N_z \times C_f}$ to construct the rendered feature map:
\begin{small}
\begin{equation}
    \setlength\abovedisplayskip{6pt}
    \setlength\belowdisplayskip{6pt}
    \hat{F}_t (r) = \sum_{i=1}^{N_s} w_{t,i} V_f [x + \text{MLP}_{\theta_d} (x, t)]
    \label{eq7}
\end{equation}
\end{small}
where $w_{t,i}$ is calculated from Eq. \ref{eq5}.
$\hat{F}_t \in \mathbb{R}^{N_h \times N_w \times C_f}$ is the rendered feature map and $C_f$ is the feature dimension.
Note that the 2D feature map is directly rendered without using any other MLP modules to aggregate the multi-view consistency in the canonical 3D space.


Aiming to align the latent feature space of our canonical feature volume and a pre-trained autoencoder $\phi$ \cite{simonyan2014very,li2019learning} pre-trained on large-scale 2D image datasets, we further utilize the following loss functions for training:
\begin{small}
\begin{equation}
    \setlength\abovedisplayskip{6pt}
    \setlength\belowdisplayskip{6pt}
    L_{\text{fea}} = \| \hat{F_t} - \phi_{\text{enc}} (I_t) \|_2^2 + \| \phi_{\text{dec}} (\hat{F}_t) - I_t \|_2^2
    \label{eq8}
\end{equation}
\end{small}
where $\hat{F}_t$ is the rendered 2D feature map of all pixels/rays.
$\phi_{\text{enc}}$ and $\phi_{\text{dec}}$ are respectively the encoder and decoder networks of the pre-trained autoencoder \cite{simonyan2014very}.

To reduce the huge computation and memory cost of volume-based rendering, we further apply the vector-matrix tensor decomposition \cite{chen2022tensorf} to factorize the tensors into compact vectors and matrices.

\subsection{Canonical Style Transformation}
\label{sec:method:canonical_style_transformation}

The combination of the canonical feature volume in Sec. \ref{sec:method:canonical_feature_distillation} and the deformation network in Sec. \ref{sec:method:preliminary} can provide a compact representation of the 4D feature space.
In this section, our goal is to find a linear style transformation matrix that is consistent on the 4D feature space. 
A naive solution is to apply the adaptive style normalization technique proposed by StyleRF \cite{liu2023stylerf} to the 3D feature volume at each frame.
However, the sampled points from the canonical space are exactly different at different times in a 4D scene, leading to a changing per-frame 3D feature volume.
Since the deferred style transformation in StyleRF depends on the mean and variance information of the changing per-frame volume, it inevitably results in an unexpected variation in the style transformation toward the whole scene thus deteriorating the temporal consistency.
Consequently, we propose canonical style transformation to learn a universal whitening/coloring transformation matrix from the 4D content scene and the 2D style image.

\subsubsection{Universal Transformation on Canonical Space}
\label{sec:method:canonical_style_transformation:universal}

Denote that $F_s \in \mathbb{R}^{C_f \times N_h N_w}$ is the flattened feature map extracted by the pre-trained encoder $\phi_{\text{enc}}$ from style image $I_s$.
The canonical feature volume $V_f$ is reshaped to a matrix form $F_c = \pi(V_f) \in \mathbb{R}^{C_f \times N_x N_y N_z}$, where $\pi (\cdot)$ is the reshaping function.

We aim to find the optimal linear transformation matrix $T \in \mathbb{R}^{C_f \times C_f}$ that can transform the latent space of the canonical feature volume $V_f$ to the feature space of $F_s$.
The style transfer problem \cite{gatys2016image} can be converted to an optimization problem to minimize the Gram matrix between the content and style:
\begin{small}
\begin{equation}
    \setlength\abovedisplayskip{6pt}
    \setlength\belowdisplayskip{6pt}
    T^* = \arg \min_{T} \frac{1}{C_f} \| \overline{F}_{cs} \overline{F}_{cs}^T - \overline{F}_s \overline{F}_s^T \|_F^2, s.t. \:\: \overline{F}_{cs} = T \overline{F}_c
    \label{eq9}
\end{equation}
\end{small}
where $\overline{F}_c = F_c - \mu (F_c)$ and $\overline{F}_s = F_s - \mu(F_s)$.
$\mu (\cdot)$ calculates the mean value of the input and $T^* \in \mathbb{R}^{C_f \times C_f}$.


The optimal transformation matrix $T$ can be obtained when:
\begin{small}
\begin{equation}
    \setlength\abovedisplayskip{6pt}
    \setlength\belowdisplayskip{6pt}
    T^* \overline{F}_c {\overline{F}_c}^T {T^*}^T = \overline{F}_s {\overline{F}_s}^T
    \label{eq10}
\end{equation}
\end{small}
By applying singular value decomposition (SVD) to the covariance matrix, we can get: $\overline{F}_c {\overline{F}_c}^T = W_c \Sigma_c {W_c}^T$, $\overline{F}_s {\overline{F}_s}^T = W_s \Sigma_s {W_s}^T$.
Substitute them in Eq. \ref{eq10}:
\begin{small}
\begin{equation}
    \setlength\abovedisplayskip{6pt}
    \setlength\belowdisplayskip{6pt}
    T^* W_c \Sigma_c {W_c}^T {T^*}^T = W_s \Sigma_s {W_s}^T
    \label{eq11}
\end{equation}
\end{small}
Then it is easy to get the closed-form solution of $T^*$:
\begin{small}
\begin{equation}
    \setlength\abovedisplayskip{6pt}
    \setlength\belowdisplayskip{6pt}
    T^* = T_s T_c = \left( W_s \Sigma_s^{\frac{1}{2}} {W_s}^T \right) \left( W_c \Sigma_c^{-\frac{1}{2}} {W_c}^T \right)
    \label{eq12}
\end{equation}
\end{small}
\noindent\textbf{3D Whitening Transform.}
Based on Eq. \ref{eq12}, we can obtain the whitened 3D feature volume $\hat{F}_c \in \mathbb{R}^{C_f \times N_xN_yN_z}$:
\begin{small}
\begin{equation}
    \setlength\abovedisplayskip{6pt}
    \setlength\belowdisplayskip{6pt}
    \tilde{F}_c = T_c \overline{F}_c = W_c \Sigma_c^{-\frac{1}{2}} {W_c}^T \overline{F}_c
    \label{eq13}
\end{equation}
\end{small}
where $\tilde{F}_c$ satisfies the constraints of $\tilde{F}_c {\tilde{F}_c}^T = I$, which means it is an uncorrelated feature volume without any style.

\noindent\textbf{3D Coloring Transform.}
Based on Eq. \ref{eq12}, we can obtain $\overline{F}_{cs} \in \mathbb{R}^{C_f \times N_xN_yN_z}$ which contains the target style satisfying the constraints: $\overline{F}_{cs} {\overline{F}_{cs}}^T = \overline{F}_s {\overline{F}_s}^T$.
\begin{small}
\begin{equation}
    \setlength\abovedisplayskip{6pt}
    \setlength\belowdisplayskip{6pt}
    \overline{F}_{cs} = T_s \tilde{F}_c = W_s \Sigma_s^{\frac{1}{2}} {W_s}^T \tilde{F}_c
    \label{eq14}
\end{equation}
\end{small}
Then we re-center the feature volume with the mean value of the style: $F_{cs} = \overline{F}_{cs} + \mu (F_s)$.

\noindent\textbf{Feature Rendering.}
Finally, we can render the transformed feature volume to 2D feature map via volume rendering in Eq. \ref{eq5}.
\begin{small}
\begin{equation}
    \setlength\abovedisplayskip{6pt}
    \setlength\belowdisplayskip{6pt}
    \hat{F}_{t,cs} (r) = \sum_{i=1}^{N_s} w_{t,i} \pi^{-1} (F_{cs}) [\tilde{x}_t]
    \label{eq15}
\end{equation}
\end{small}
where $\pi^{-1}$ is the inverse reshaping function that converts matrix $\hat{F}_{cs}$ to a feature volume.
Note that we only need to compute the universal transformation for one time and we can render the novel views at different time $t$ with the weight $w_{t,i}$ calculated by Eq. \ref{eq5}.
The rendered feature map is further fed to the pre-trained decoder $\phi_{\text{dec}}$ to reconstruct the stylized RGB image:
\begin{small}
\begin{equation}
    \setlength\abovedisplayskip{6pt}
    \setlength\belowdisplayskip{6pt}
    I_{t,cs} = \phi_{\text{dec}} (\hat{F}_{t,cs})
    \label{eq16}
\end{equation}
\end{small}


\subsubsection{Learning Transformation from Data}
\label{sec:method:canonical_style_transformation:learning}

Solving the closed-form solution from Eq. \ref{eq9} to \ref{eq12} creates a huge burden in memory and computation, especially the SVD operations on the whole 3D feature volume.
Hence, we aim to design an alternative way to estimate these linear transformation matrices with feed-forward networks in a data-driven manner instead of calculating the closed-form solution.

The 3D style transformations can be rewritten in one equation:
\begin{small}
\begin{equation}
    \setlength\abovedisplayskip{6pt}
    \setlength\belowdisplayskip{6pt}
    \hat{F}_{t,cs} (r) = \sum_{i=1}^{N_s} w_{t,i} \left( \pi^{-1} \left(T_s T_c (\pi(V_f) - \mu_f) + \mu_s \right) [\tilde{x}_t] \right)
    \label{eq17}
\end{equation}
\end{small}
where $T_s \in \mathbb{R}^{C_f \times C_f}$ is the coloring matrix and $T_s \in \mathbb{R}^{C_f \times C_f}$ is the whitening matrix.
$\mu_f = \mu(\pi(V_f))$ is the mean value of the canonical feature volume $V_f$.
$\mu_s = \mu(F_s)$ is the mean value of the style feature.
We can further organize the formula as follows:
\begin{small}
\begin{equation}
    \setlength\abovedisplayskip{6pt}
    \setlength\belowdisplayskip{6pt}
    \hat{F}_{t,cs} (r) = \pi^{-1} \left(T_s T_c \left(\pi \left( \sum_{i=1}^{N_s} w_{t,i} V_f [\tilde{x}_t] \right) - \mu_f \right) + \mu_s\right) 
    \label{eq18}
\end{equation}
\end{small}
Substituting the corresponding terms of Eq. \ref{eq18} with Eq. \ref{eq8}:
\begin{small}
\begin{equation}
    \setlength\abovedisplayskip{6pt}
    \setlength\belowdisplayskip{6pt}
    \hat{F}_{t,cs} (r) = \pi^{-1} \left(T_s T_c \left(\pi \left( \hat{F}_t (r) \right) - \mu_f \right) + \mu_s \right)
    \label{eq19}
\end{equation}
\end{small}
From the above equation, we can introduce the key design of the proposed canonical style transformation according to the following findings:

1) \textbf{$\mu_f$ is fixed value because the canonical feature volume $V_f$ is fixed.}
Hence we can calculate $\mu_f$ in advance to retrench the computation.

2) \textbf{$T_c$ is only conditioned by the content feature volume $V_f$.}
We adopt a lightweight 3D CNN $\psi_{\text{3D}}$ to process the content feature volume:
\begin{small}
\begin{equation}
    \setlength\abovedisplayskip{6pt}
    \setlength\belowdisplayskip{6pt}
    V_z = \psi_{\text{3D}} (V_f - \mu_f)
    \label{eq20}
\end{equation}
\end{small}
where $V_z \in \mathbb{R}^{N_x \times N_y \times N_z \times C_z}$ is the output volume.
$C_z$ is the output channel dimension which is smaller than the original feature dimension $C_f$.
Then, we can compute the covariance matrix of $V_z$ and adopt a MLP parameterized by $\theta_{cz}$ to predict the whitening matrix:
\begin{small}
\begin{equation}
    \setlength\abovedisplayskip{6pt}
    \setlength\belowdisplayskip{6pt}
    T_c =  \text{MLP}_{\theta_{cz}} (\text{cov} (\pi(V_z)))
    \label{eq21}
\end{equation}
\end{small}
where $\text{cov} (\pi(V_z)) \in \mathbb{R}^{C_z \times C_z}$ is the covariance matrix of $\pi(V_z)$.
$\pi$ is the reshaping function to flatten a volume to a matrix.
The MLP maps the squeezed feature dimension from $C_z$ to the original $C_f$ in $\hat{T}_c \in \mathbb{R}^{C_f \times C_f}$.

3) \textbf{$T_s$ is only conditioned by the style feature $F_s$.}
We can use a lightweight 2D CNN $\psi_{\text{2D}}$ to compress the dimension of feature map $F_s$:
\begin{small}
\begin{equation}
    \setlength\abovedisplayskip{6pt}
    \setlength\belowdisplayskip{6pt}
    F_z = \psi_{\text{2D}} (F_s - \mu_s)
    \label{eq22}
\end{equation}
\end{small}
where $F_z$ has a channel dimension of $C_z$.
The coloring matrix $\hat{T}_s \in \mathbb{R}^{C_f \times C_f}$ can be estimated with another MLP paramterized by $\theta_{sz}$ mapping the covariance matrix of $F_z$ from $C_z$ to the original dimension of $C_f$:
\begin{small}
\begin{equation}
    \setlength\abovedisplayskip{6pt}
    \setlength\belowdisplayskip{6pt}
    T_s = \text{MLP}_{\theta_{sz}} (\text{cov} (\pi ( F_z )))
    \label{eq23}
\end{equation}
\end{small}

4) \textbf{Eq. \ref{eq19} is an equivalent form of Eq. \ref{eq17}.}
The matrix transformations in Eq. \ref{eq17} are applied on the whole 3D feature volume which is pretty computation-intensive.
Whereas, the rendered feature map $\hat{F}_t$ in Eq. \ref{eq19} can be computed in advance, and the computation can be significantly reduced because only 2D feature map is required.

In summary, we replace the original closed-form matrix transformations in Eq. \ref{eq19} with the predictions of feed-forward networks.
The equivalent form in Eq. \ref{eq19} can lower the computation complexity of the volume-based operations from $\mathcal{O}(n^3)$ in Eq. \ref{eq17}  to $\mathcal{O}(n^2)$.
It allows us to conduct a universal style transformation towards the canonical volume and consequently preserve the cross-time consistency.
For training, the predicted feature map $\hat{F}_{t,cs}$ can be supervised by traditional style loss \cite{gatys2016image}:
\begin{small}
\begin{equation}
\begin{aligned}
    \setlength\abovedisplayskip{6pt}
    \setlength\belowdisplayskip{6pt}
    L_{\text{sty}} = \lambda_{\text{con}} * \| \phi_{\text{enc}} (\phi_{\text{dec}}(\hat{F}_{t,cs})) - \phi_{\text{enc}} (I_t) \|_2^2 + \\
      \lambda_{\text{sty}} * \| \text{Gram} (\hat{F}_{t,cs}) - \text{Gram} (\phi_{\text{enc}} (I_s)) \|_2^2
    \label{eq24}
\end{aligned}
\end{equation}
\end{small}
where the former term is the content constraint and the latter one is the style constraint.
$\text{Gram} (\cdot)$ calculates the Gram matrix \cite{gatys2016image} of the input.
In default, $\lambda_{\text{con}} = 1$ and $\lambda_{\text{sty}} = 10$.

\begin{figure*}
    \centering
    \includegraphics[width=\textwidth]{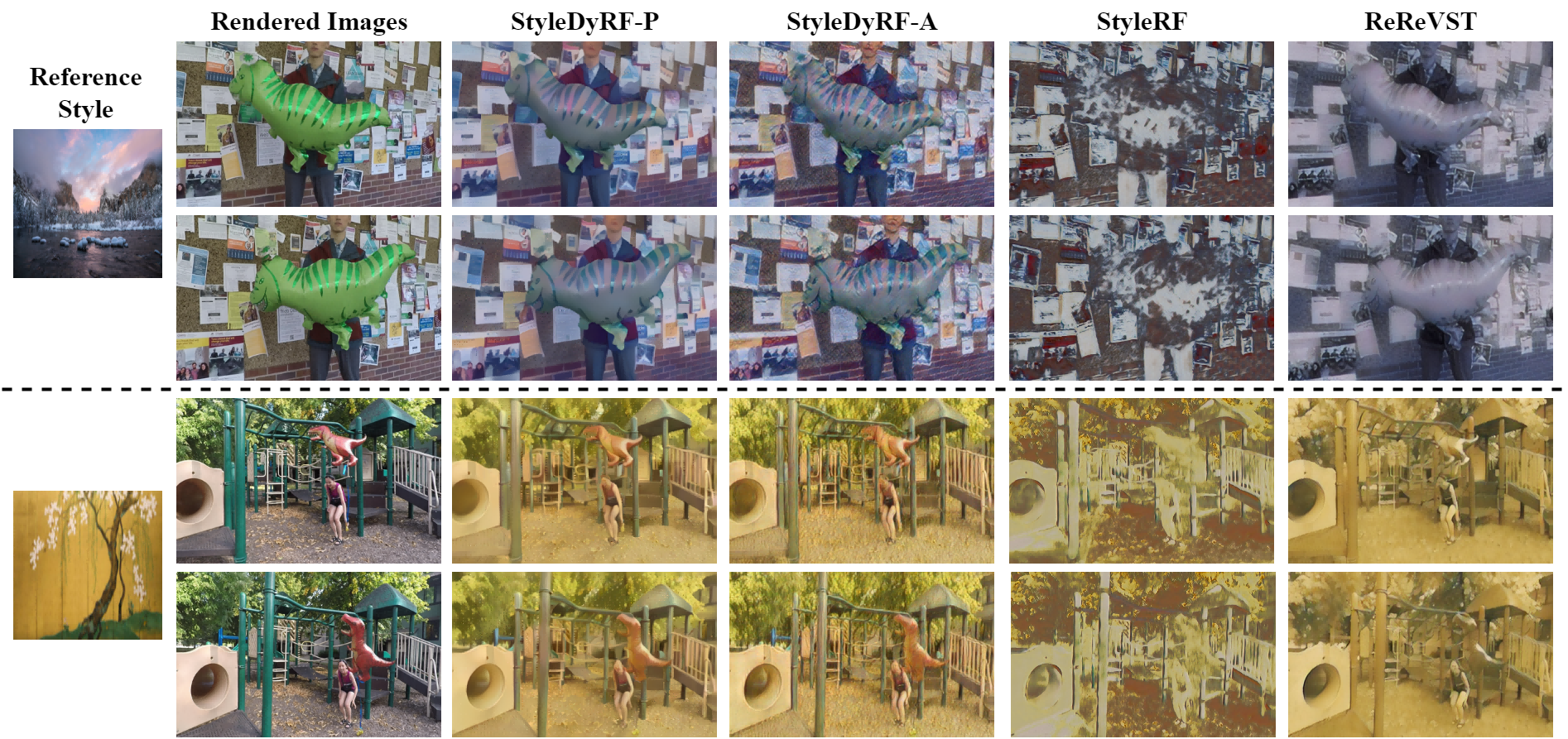}
    \vspace{-0.6cm}
    \caption{\textbf{Comparison of our StyleDyRF with other methods on Nvidia Dataset.} Our StyleDyRF produces better stylized novel views with temporal and multi-view consistency in the provided samples of dynamic scenes.}
    \vspace{-0.3cm}
    \label{fig:qualitative_comparison}
\end{figure*}

\begin{figure*}
    \centering
    \includegraphics[width=\textwidth]{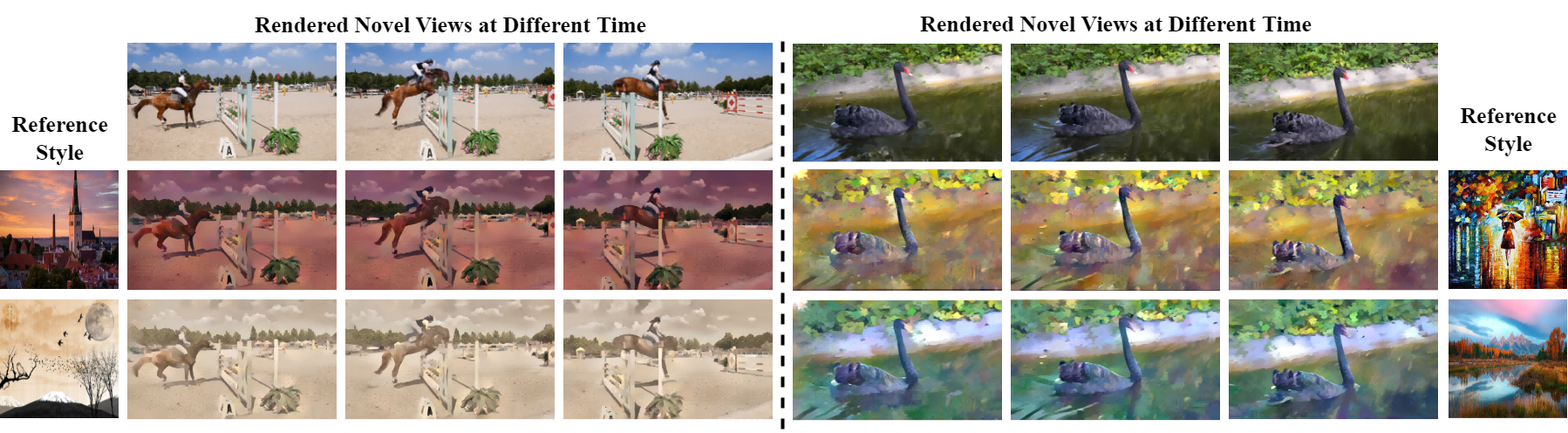}
    \vspace{-0.6cm}
    \caption{\textbf{Qualitative results of our StyleDyRF on DAVIS dataset.} Our StyleDyRF can render stylized novel views with realistic quality and 4D consistency at different times.}
    \vspace{-0.3cm}
    \label{fig:qualitative_davis}
\end{figure*}

\subsubsection{Learning Photo-realistic Style Transfer}
\label{sec:method:canonical_style_transformation:photorealistic}

Photo-realistic 4D style transfer is another important application that aims to render novel views without structural distortions.
It is proved by prior works \cite{li2019learning} that the stylized results may contain apparent artifacts because of the distortion caused by the pre-trained deep auto-encoder \cite{simonyan2014very}.
Our StyleDyRF can be simply extended to a photo-realistic style transfer framework by applying an additional post-processing network.
Inspired by \cite{li2019learning}, we can apply a convolution spatial propagation network (CSPN) \cite{cheng2019learning} as a filter network for distortions.
The network takes the rendered novel views $\hat{C}_t$ by Eq. \ref{eq4} and \ref{eq5} as the conditional input and filters the stylized output $\hat{F}_{t,cs}$ of Eq. \ref{eq19}.
\begin{small}
\begin{equation}
    \setlength\abovedisplayskip{6pt}
    \setlength\belowdisplayskip{6pt}
    I_{t,p} = \text{CSPN} (\phi_{\text{dec}} (\hat{F}_{t,cs}), \text{Whiten} (\hat{C}_t))
\end{equation}
\end{small}
where the rendered RGB image $\hat{C}_t$ at novels view is firstly whitened to obtain an uncorrelated map in the RGB space via $\text{Whiten} (\cdot)$ function which is a degraded 2D version of Eq. \ref{eq13}.
To train the CSPN network, we apply the following loss for training:

\begin{small}
\begin{equation}
    \setlength\abovedisplayskip{6pt}
    \setlength\belowdisplayskip{6pt}
    L_{fil} = \| \phi_{\text{enc}} (I_{t,p}) - \phi_{\text{enc}}(I_t) \|_2^2 + \| \nabla (I_{t,p}) - \nabla (I_t) \|_2^2
\end{equation}
\end{small}
where the former term is the content constraints and the latter one is the gradient regularization loss on the image.




\section{Experiments}
\label{sec:exp}


We conduct qualitative experiments (Sec. \ref{sec:exp:qualitative}) and quantitative experiments (Sec. \ref{sec:exp:quantitative}) to evaluate our method.
Furthermore, we evaluate the proposed method with ablation studies (Sec. \ref{sec:exp:ablation}) on the impact of the ingredients in StyleDyRF and a new application of multi-style interpolation (Sec. \ref{sec:exp:interpolation}).
We evaluate our method on 2 public datasets: Nvidia dataset \cite{li2021neural} and DAVIS dataset \cite{perazzi2016benchmark}.
Following StyleRF \cite{liu2023stylerf}, we select WikiArt \cite{wikiart} as our style dataset.
The Dynamic NeRF-pretraining stage requires 10,000 iterations, while each of the remaining 2 training stages requires 2,500 iterations.
The whole training process takes about 1 day on a single NVIDIA V100.
Further implementation details are provided in the supplementary materials.







\subsection{Qualitative Results}
\label{sec:exp:qualitative}

Fig. \ref{fig:qualitative_comparison} shows the qualitative comparisons on the Nvidia dataset between our StyleDyRF and other methods including: the state-of-the-art NeRF-based style transfer method StyleRF \cite{liu2023stylerf}, and a video-based style transfer method ReReVST \cite{wang2020consistent}.
In the figure, StyleDyRF-A is the artistic style transfer model of our proposed method, and StyleDyRF-P has an additional post-processing propagation module introduced in Sec. \ref{sec:method:canonical_style_transformation:photorealistic}.
From the rendered novel views in the figure, we can find that our StyleDyRF can generate novel views with better stylized novel views compared with other methods.
StyleRF fails to handle the dynamic objects, resulting in vague artifacts in dynamic regions.
Many detailed texture information is dropped because of the excessive regularization in ReReVST.
We also provide the qualitative results on the DAVIS dataset in Fig. \ref{fig:qualitative_comparison}.
It shows that our method can robustly generalize to new style patterns in a zero-shot manner meantime preserving multi-view consistency and cross-time consistency during stylization.


\subsection{Quantitative Results}
\label{sec:exp:quantitative}

\begin{table*}[t]
\centering
\small
\resizebox{\linewidth}{!}{
\begin{tabular}{l|cc|cc|cc|cc|cc|cc|cc|cc}
\hline
\multirow{2}{*}{Method} & \multicolumn{2}{c|}{\emph{Balloon1}} & \multicolumn{2}{c|}{\emph{Balloon2}} & \multicolumn{2}{c|}{\emph{Jumping}} & \multicolumn{2}{c|}{\emph{Skating}} & \multicolumn{2}{c|}{\emph{Playground}} & \multicolumn{2}{c|}{\emph{Truck}} & \multicolumn{2}{c|}{\emph{Umbrella}} & \multicolumn{2}{c}{\emph{\textbf{Mean}}} \\
                        & \textbf{LPIPS}     & \textbf{RMSE}    & \textbf{LPIPS}     & \textbf{RMSE}    & \textbf{LPIPS}    & \textbf{RMSE}    & \textbf{LPIPS}    & \textbf{RMSE}    & \textbf{LPIPS}      & \textbf{RMSE}     & \textbf{LPIPS}   & \textbf{RMSE}   & \textbf{LPIPS}     & \textbf{RMSE}     & \textbf{LPIPS}    & \textbf{RMSE}    \\ \hline \hline
WCT \cite{li2017universal}                    & 0.347         & 0.201        & 0.351         & 0.186        & 0.341        & 0.204        & 0.375        & 0.193        & 0.303          & 0.191         & 0.336       & 0.163       & 0.360         & 0.204         & 0.345        & 0.192        \\
LinearWCT \cite{li2019learning}             & 0.219 & 0.101 & 0.148 & 0.075 & 0.195 & 0.095 & 0.163 & 0.086 & 0.201 & 0.092 & 0.141 & 0.084 & 0.164 & 0.091 & 0.176 & 0.089 \\
AdaIN \cite{huang2017arbitrary}                  & 0.253         & 0.073        & 0.170         & 0.057        & 0.208        & 0.072        & 0.198        & 0.072        & 0.220          & 0.070         & 0.183       & 0.068       & 0.207         & 0.072         & 0.206        & 0.069        \\
MCCNet  \cite{deng2021arbitrary}                & 0.229         & 0.088        & 0.136         & 0.059        & 0.189        & 0.080        & 0.162        & 0.072        & 0.199          & 0.075         & 0.137       & 0.071       & 0.160         & 0.076         & 0.173        & 0.074        \\
ReReVST \cite{wang2020consistent}                & 0.179         & 0.082        & 0.102         & 0.055        & 0.149        & 0.068        & 0.125        & 0.067        & 0.156          & 0.065         & 0.103       & 0.061       & 0.125         & 0.071         & 0.134        & 0.067        \\ \hline
StyleRF \cite{liu2023stylerf}                & 0.159         & 0.105        & 0.114         & 0.085        & 0.192        & 0.106        & 0.130        & 0.081        & 0.130          & 0.079         & 0.137       & 0.085       & 0.120         & 0.083        & 0.140        & 0.089        \\
StyleDyRF-A              & 0.094         & 0.060        & 0.095         & 0.047        & 0.086        & 0.055        & 0.092        & 0.047        & \textbf{0.096}          & 0.060         & 0.103       & 0.053       & 0.100         & 0.051         & 0.095        & 0.053        \\
StyleDyRF-P                 & \textbf{0.086}         & \textbf{0.054}        & \textbf{0.089}         & \textbf{0.045}        & \textbf{0.073}        & \textbf{0.050}        & \textbf{0.068}        & \textbf{0.044}        & {0.097}          & \textbf{0.060}         & \textbf{0.070}       & \textbf{0.040}       & \textbf{0.079}         & \textbf{0.040}         & \textbf{0.080}        & \textbf{0.048}        \\ \hline
\end{tabular}}
\vspace{-0.3cm}
\caption{\textbf{Quantitative comparisons on short-range consistency.} We compare the consistency scores of LPIPS ($\downarrow$) and RMSE ($\downarrow$) between stylized images at 2 adjacent novel views to evaluate the short-range consistency.}
\vspace{-0.3cm}
\label{tab:quantitative_shortrange}
\end{table*}

\begin{table*}[t]
\centering
\small
\resizebox{\linewidth}{!}{
\begin{tabular}{l|cc|cc|cc|cc|cc|cc|cc|cc}
\hline
\multirow{2}{*}{Method} & \multicolumn{2}{c|}{\emph{Balloon1}} & \multicolumn{2}{c|}{\emph{Balloon2}} & \multicolumn{2}{c|}{\emph{Jumping}} & \multicolumn{2}{c|}{\emph{Skating}} & \multicolumn{2}{c|}{\emph{Playground}} & \multicolumn{2}{c|}{\emph{Truck}} & \multicolumn{2}{c|}{\emph{Umbrella}} & \multicolumn{2}{c}{\emph{\textbf{Mean}}} \\
                        & \textbf{LPIPS}     & \textbf{RMSE}    & \textbf{LPIPS}     & \textbf{RMSE}    & \textbf{LPIPS}    & \textbf{RMSE}    & \textbf{LPIPS}    & \textbf{RMSE}    & \textbf{LPIPS}      & \textbf{RMSE}     & \textbf{LPIPS}   & \textbf{RMSE}   & \textbf{LPIPS}     & \textbf{RMSE}     & \textbf{LPIPS}    & \textbf{RMSE}    \\ \hline \hline
WCT \cite{li2017universal}                    & 0.349         & 0.200        & 0.361         & 0.200        & 0.339        & 0.212        & 0.375        & 0.204        & 0.289          & 0.202         & 0.337       & 0.174       & 0.373         & 0.221        & 0.346        & 0.202        \\
LinearWCT \cite{li2019learning}            & 0.223 & 0.094 & 0.183 & 0.090 & 0.260 & 0.120 & 0.206 & 0.097 & 0.248 & 0.106 & 0.170 & 0.085 & 0.229 & 0.103 & 0.217 & 0.099 \\
AdaIN \cite{huang2017arbitrary}                  & 0.236         & 0.067        & 0.195         & 0.065        & 0.250        & 0.078        & 0.226        & 0.073        & 0.238          & 0.074         & 0.199       & 0.061       & 0.253         & 0.080        & 0.228        & 0.071        \\
MCCNet  \cite{deng2021arbitrary}                & 0.222         & 0.082        & 0.171         & 0.074        & 0.250        & 0.098        & 0.209        & 0.081        & 0.248          & 0.089         & 0.162       & 0.068       & 0.220         & 0.095        & 0.212        & 0.084        \\
ReReVST  \cite{wang2020consistent}               & 0.181         & 0.078        & 0.133         & 0.061        & 0.218        & 0.096        & 0.181        & 0.088        & 0.207          & 0.082         & 0.124       & 0.053       & 0.185         & 0.084        & 0.176        & 0.077        \\ \hline
StyleRF  \cite{liu2023stylerf}               & 0.198         & 0.119        & 0.133         & 0.068        & 0.279        & 0.141        & 0.199        & 0.109        & 0.193          & 0.098         & 0.194       & 0.096       & 0.172         & 0.089        & 0.195        & 0.103        \\
StyleDyRF-A                  & 0.118         & 0.069        & 0.113         & 0.058        & 0.107        & 0.070        & 0.103        & 0.061        & 0.114          & 0.074         & 0.117       & 0.058       & 0.114         & 0.061        & 0.112        & 0.065        \\
StyleDyRF-P                  & \textbf{0.112}         & \textbf{0.063}        & \textbf{0.111}         & \textbf{0.058}        & \textbf{0.101}        & \textbf{0.068}        & \textbf{0.086}        & \textbf{0.060}        & \textbf{0.113}          & \textbf{0.073}         & \textbf{0.084}       & \textbf{0.046}       & \textbf{0.097}         & \textbf{0.050}        & \textbf{0.101}        & \textbf{0.060}        \\ \hline
\end{tabular}}
\vspace{-0.3cm}
\caption{\textbf{Quantitative comparisons on long-range consistency.} We compare the consistency scores of LPIPS ($\downarrow$) and RMSE ($\downarrow$) between stylized images at 2 novel views which are far away in time sequence to evaluate the short-range consistency.}
\vspace{-0.3cm}
\label{tab:quantitative_longrange}
\end{table*}

Similar to the 3D style transfer task, there are few metrics for quantitative evaluations of stylization quality in our 4D style transfer task.
Previous works \cite{chiang2022stylizing,liu2023stylerf} attempt to evaluate the multi-view consistency of the generated novel views of different methods.
In our experiments, the evaluation of cross-time consistency is also required to be evaluated in our benchmark besides the multi-view consistency.
Consequently, the evaluation also includes the variation along the time sequence.
To compute the quantitative metrics, we can warp one view to another view according to the optical flow \cite{teed2020raft} using softmax splitting \cite{niklaus2020softmax} and evaluate the LPIPS score \cite{zhang2018unreasonable} and RMSE score to measure the consistency after stylization.
Following prior works \cite{liu2023stylerf,chiang2022stylizing}, we evaluate the consistency in two different ways: 1) short-range consistency between adjacent views, i.e. $t$ and $t+1$ frames in the testing video; 2) long-range consistency between far-way novel views, i.e. $t$ and $t+T/3$ frames in the test video with $T$ frames in total.
The quantitative comparison of short-range consistency and long-range consistency on the Nvidia dataset are respectively provided in Tab. \ref{tab:quantitative_shortrange} and \ref{tab:quantitative_longrange}.
The comparison includes: 1) Image-based style transfer techniques: WCT \cite{li2017universal}, LinearWCT \cite{li2019learning}, AdaIN \cite{huang2017arbitrary}; 2) Video-based style transfer methods: MCCNet  \cite{deng2021arbitrary}, ReReVST  \cite{wang2020consistent}; 3) NeRF-based style transfer methods: StyleRF \cite{liu2023stylerf}.
From the tables, it can be seen that our StyleDyRF significantly outperforms other approaches in both short-range and long-range consistency.
It demonstrates that our StyleDyRF can effectively stylize the novel's views with consistency in dynamic scenes.


\subsection{Ablation Studies}
\label{sec:exp:ablation}

\begin{figure}
    \centering
    \includegraphics[width=0.8\linewidth]{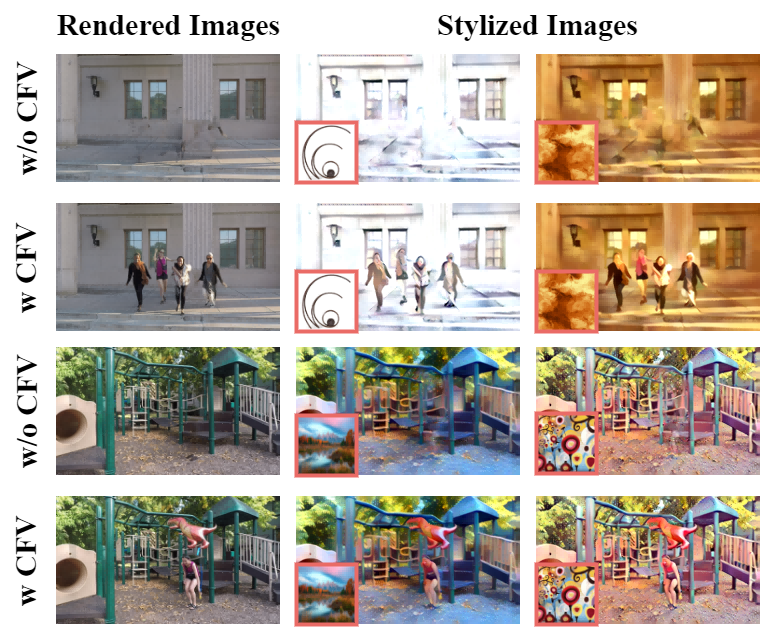}
    \vspace{-0.3cm}
    \caption{\textbf{Ablation results of our proposed canonical feature volume.} CFV can effectively model dynamic objects in 4D scenes, providing coherent stylization with the whole scene.}
    \vspace{-0.3cm}
    \label{fig:ablation_cfv}
\end{figure}


\begin{figure}
    \centering
    \includegraphics[width=0.8\linewidth]{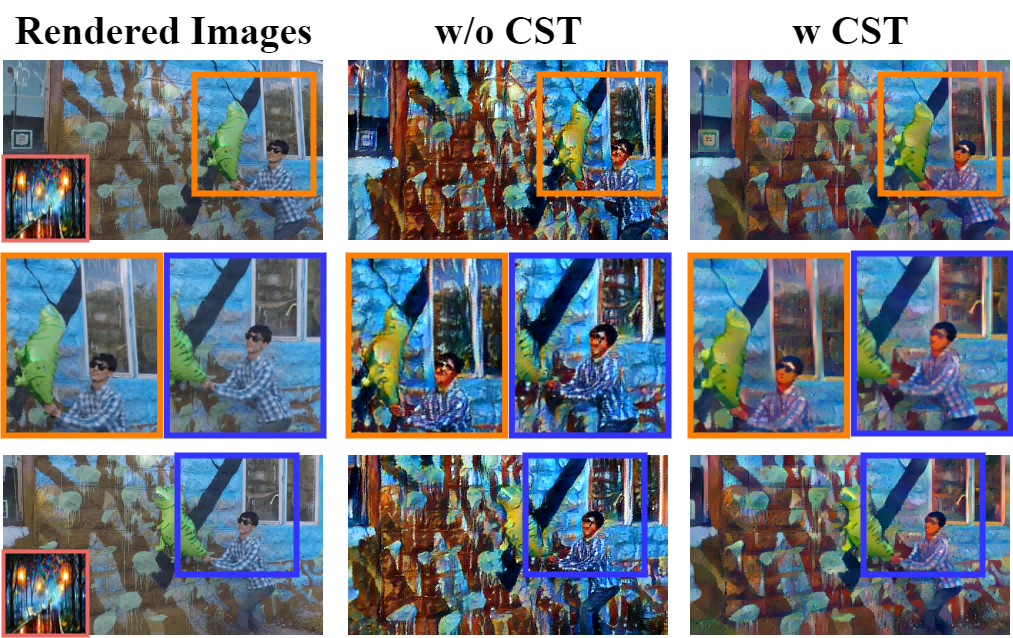}
    \vspace{-0.3cm}
    \caption{\textbf{Ablation results of our proposed canonical style transformation.} CST can better preserve the global consistency of the 4D scene in the stylized outputs.}
    \vspace{-0.3cm}
    \label{fig:ablation_cst}
\end{figure}


\noindent\textbf{Ablation Study on Canonical Feature Volume.} We first evaluate the design of Canonical Feature Volume (CFV) in Fig. \ref{fig:ablation_cfv}.
In the figure, the baseline "w/o CFV" removes the CFV in our StyleDyRF, and it only utilizes a static 3D feature volume during training.
We can find that the dynamic objects are invisible in the stylized results of "w/o CFV".
The involvement of CFV can effectively model the dynamic objects in 4D scenes meantime conducting coherent style transformation with the whole scene.

\noindent\textbf{Ablation Study on Canonical Style Transformation.} We also evaluate the design of Canonical Style Transformation (CST) discussed in Sec. \ref{sec:method:canonical_style_transformation} in Fig. \ref{fig:ablation_cst}.
The 1st and 3rd row shows two different novel views, and the 2nd row shows the zoomed region.
For comparison, the baseline "w/o CST" first renders the RGB images and feeds them to the traditional style transfer method \cite{li2019learning}.
From the comparison, our CST (3-rd column) produces much less distortions than the baseline.
It demonstrates the superiority of the proposed CST towards cross-time consistency and multi-view consistency.

\begin{figure}
    \centering
    \includegraphics[width=0.8\linewidth]{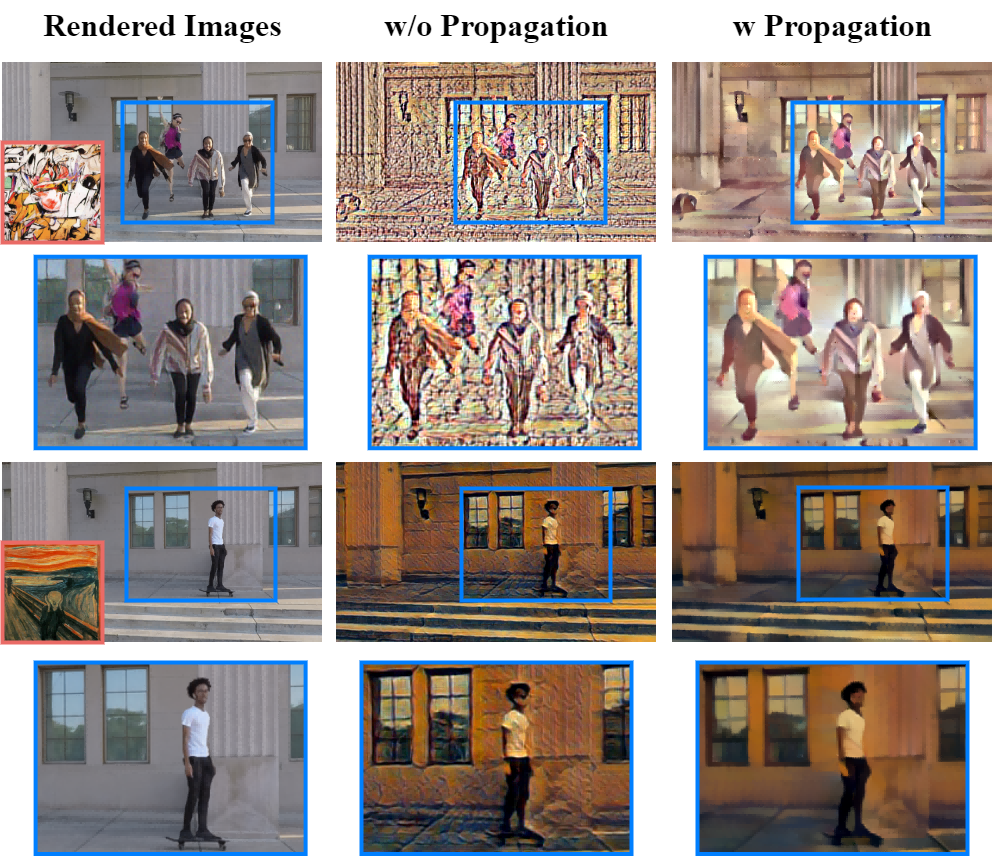}
    \vspace{-0.3cm}
    \caption{\textbf{Ablation results of the spatical propagation module.} The propagation module in StyleDyRF can effectively filter the structure distortions \cite{li2019learning} caused by the pre-trained autoencoder.}
    \vspace{-0.3cm}
    \label{fig:ablation_spn}
\end{figure}

\noindent\textbf{Ablation Study on Propagation Module.}
We further evaluate the effect of the post-processing CSPN module discussed in Sec. \ref{sec:method:canonical_style_transformation:photorealistic} in Fig. \ref{fig:ablation_spn}.
As the figure shows, the structure distortions caused by style transformation in artistic style transfer can be effectively filtered with the well-trained CSPN module.
It can provide high-quality stylization results in the 4D photo-realistic style transfer task.

\subsection{Multi-style Interpolation}
\label{sec:exp:interpolation}

\begin{figure}
    \centering
    \includegraphics[width=0.9\linewidth]{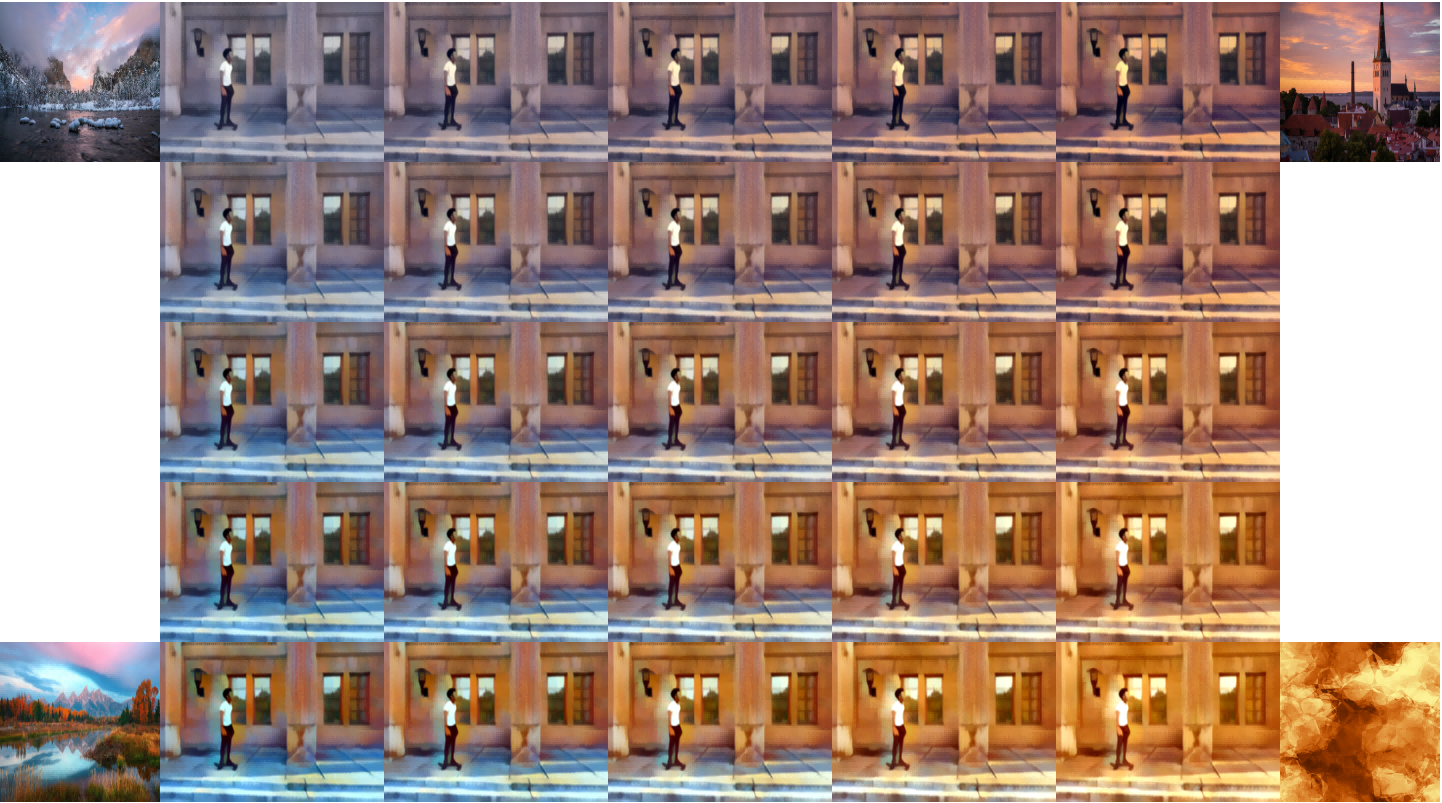}
    \vspace{-0.3cm}
    \caption{\textbf{Multiple-style interpolation.} Our StyleDyRF can smoothly interpolate between different styles via interpolating the linear transformation matrix in CST.}
    \vspace{-0.3cm}
    \label{fig:interpolation}
\end{figure}

Since the designed CST is a linear transformation on the latent feature space, our StyleDyRF can smoothly merge different style patterns by linearly interpolating the coloring matrix $F_s$ and the mean feature $\mu_s$ in Eq. \ref{eq20}.
In Fig. \ref{fig:interpolation}, we provide the interpolation results of 4 different style patterns in the same dynamic scene.
It demonstrates the expressive generalization ability of CST in the latent space and further shows the potential of generalization to unseen style patterns via multi-style interpolation.
\section{Conclusion}


In this paper, we first consider the problem of 4D style transfer that aims to transfer arbitrary style patterns to the synthesized novel views in a dynamic 4D scene.
To handle this issue, we propose StyleDyRF, a novel method that models the 4D feature space with a canonical feature volume and learns the style transformation matrix on the feature volume in a learning-based fashion.
We utilize a compact representation of canonical feature volume to model the 4D scene by deforming it with a pre-trained deformation network in dynamic NeRF.
To ensure temporal consistency, we propose a learnable fashion to approximate the style transformation matrix based on the 3D canonical feature volume and the style images.
The experimental results demonstrate that StyleDyRF can effectively render high-quality stylized novel views consistently, and also be extended to potential applications for artistic 4D scenes.
{
    \small
    \bibliographystyle{unsrt}
    \bibliography{main}
}


\end{document}